\documentclass{article}
\usepackage{ijcai20}

\usepackage{times}

\usepackage{soul}
\usepackage{url}
\usepackage[hidelinks]{hyperref}
\usepackage[utf8]{inputenc}
\usepackage[small]{caption}
\usepackage{graphicx}
\usepackage{amsmath}
\usepackage{amsthm}
\usepackage{amssymb}
\usepackage{booktabs}
\usepackage{algorithm}
\usepackage{algorithmic}
\usepackage{tabularx}
\usepackage{graphicx}
\usepackage{setspace}
\usepackage{subfigure}
\usepackage{caption}
\usepackage{multirow}
\usepackage[left]{lineno}
\usepackage{color}

\title{P-KDGAN: Progressive Knowledge Distillation with GANs \\ for One-class Novelty Detection}

\author{
Zhiwei Zhang$^{1,2}$\footnotemark[1]\and
Shifeng Chen$^1$\footnotemark[2]\And
Lei Sun$^2$ \\
\affiliations
$^1$Shenzhen Institutes of Advanced Technology, Chinese Academy of Sciences, China \\
$^2$School of Information and Electronics, Beijing Institute of Technology, China \\
\emails
\{zw.zhang3, shifeng.chen\}@siat.ac.cn,
sunlei@bit.edu.cn
}

\begin{document}
	\maketitle
	
	\begin{abstract}
		One-class novelty detection is to identify anomalous instances that do not conform to the expected normal instances. In this paper, the Generative Adversarial Networks (GANs) based on encoder-decoder-encoder pipeline are used for detection and achieve state-of-the-art performance. However, deep neural networks are too over-parameterized to deploy on resource-limited devices. Therefore, Progressive Knowledge Distillation with GANs (P-KDGAN) is proposed to learn compact and fast novelty detection networks. The P-KDGAN is a novel attempt to connect two standard GANs by the designed distillation loss for transferring knowledge from the teacher to the student. The progressive learning of knowledge distillation is a two-step approach that continuously improves the performance of the student GAN and achieves better performance than single step methods. In the first step, the student GAN learns the basic knowledge totally from the teacher via guiding of the pre-trained teacher GAN with fixed weights. In the second step, joint fine-training is adopted for the knowledgeable teacher and student GANs to further improve the performance and stability. The experimental results on CIFAR-10, MNIST, and FMNIST show that our method improves the performance of the student GAN by 2.44\%, 1.77\%, and 1.73\% when compressing the computation at ratios of 24.45:1, 311.11:1, and 700:1, respectively.
	\end{abstract}

	\renewcommand{\thefootnote}{\fnsymbol{footnote}}
	\footnotetext[1]{This work was done when Zhiwei Zhang was a research intern at Multimedia Laboratory of SIAT.}
	\footnotetext[2]{Corresponding author.}

	\section{Introduction}
	One-class novelty detection aims to identify patterns that do not belong to the normal data distribution \cite{survey:2009}. Unlike traditional classification problem, novelty detection is usually trained in an unsupervised setting where novelty data is absent. Novelty detection has a wide variety of applications such as network intrusion~\cite{intrusion:2009}, credit card fraud~\cite{fraud:2008}, medical diagnoses~\cite{AnoGAN:2017} and many more. With the advantage of deep learning, novelty detection based on generative adversarial networks (GANs) has shown state-of-the-art performance by learning the representative latent space of high-dimensional data~\cite{AnoGAN:2017,EfficientGAN:2018,OCGAN:2019}. However, deep neural networks with high computational costs and large storage prohibit their deployment to computation and memory resource limited systems.
	
	For tackling the above issue, neural network compression has been widely applied in recent years~\cite{survey:2017}. As one of the mainstream compression methods, Knowledge Distillation (KD) following a teacher-student paradigm transfers knowledge from a teacher network with higher performance to a student network. The early contributions used the outputs of the softmax layers or intermediate layers in teacher networks to improve the performance of student networks~\cite{Hinton:2015,Fitnets:2015}. In the later researches, the discriminator losses were proposed to evaluate the distinction between the distribution spaces of teacher and student networks~\cite{KDGAN:2018,Portable:2018,KTAN:2018}. To our knowledge, there is no related works on two standard GANs~\cite{GAN:2014} including two generators and two discriminators to design distillation loss for knowledge distillation. Additionally, there are rare works investigating the initialization of student networks and always random initialization is used. Our experiments demonstrate that student networks without ``knowledge'' reserve (with random initialization) do not mimic the outputs of teacher networks well. 
	
	\begin{figure*}
		\centering
		\includegraphics[scale=0.715]{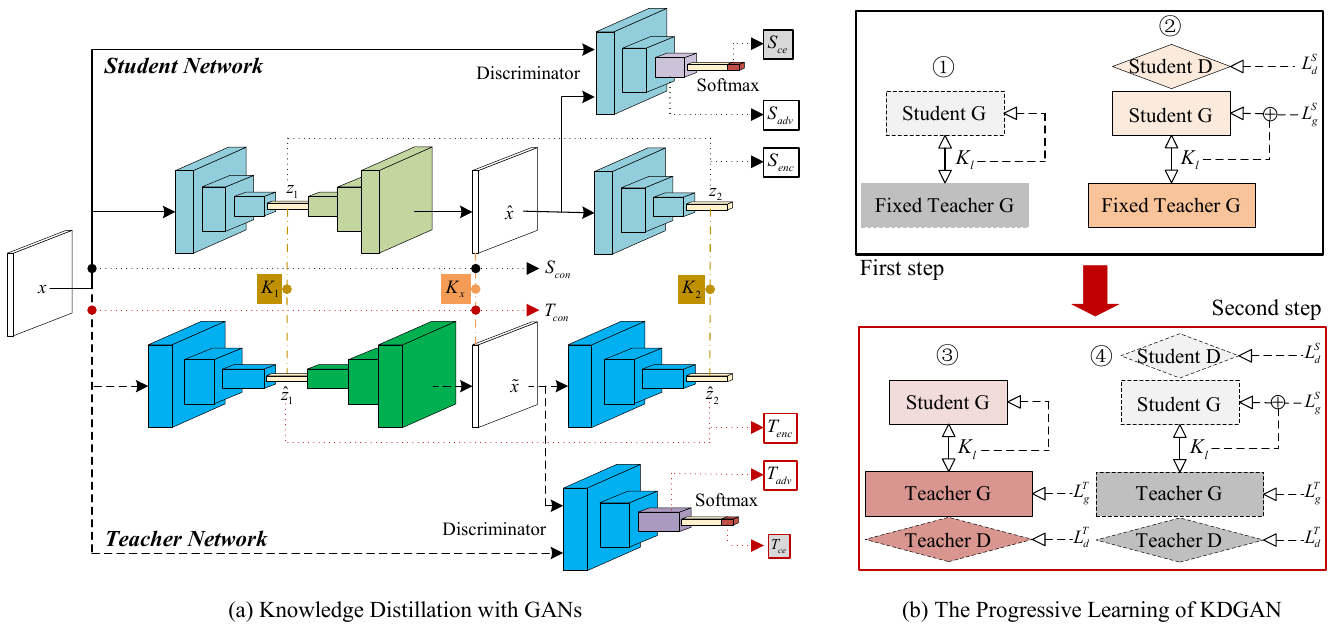}
		\caption{The flowchart of Knowledge Distillation with GANs for One-class Novelty Detection. (a) The Knowledge Distillation with Generative Adversarial Networks (KDGAN), in which the distillation losses $K_l$ ($K_l$ = $w_1K_1$ +$w_xK_x$+ $w_2K_2$) is designed for training the student GAN. (b) The two-step progressive learning of KDGAN is used to continuously improve the performance of the student GAN. KDGAN-\normalsize{\textcircled{\footnotesize{1}}}, KDGAN-\normalsize{\textcircled{\footnotesize{2}}}, KDGAN-\normalsize{\textcircled{\footnotesize{3}}} and KDGAN-\normalsize{\textcircled{\footnotesize{4}}} are four different distillation structures.}\label{fig:P-KDGAN}
	\end{figure*}
	
	In this paper, we apply GANs in the encoder-decoder-encoder structure~\cite{ganomaly:2018} for one-class novelty detection, which outperforms the state-of-the-art approaches. In order to deploy the deep neural networks in computation resources limited mobile devices, we propose the Progressive Knowledge Distillation with GANs (P-KDGAN) method to train the lightweight student network. The P-KDGAN approach improves the performance of student GAN by solving the following three problems. 
	1) How to design a distillation loss to measure the similarity of intermediate representations learned from the teacher GAN and the student GAN? As is shown in the student GAN of Figure \ref{fig:P-KDGAN}(a), the generator based on encoder-decoder-encoder pipeline can generate two latent vectors $z_1$, $z_2$ and a reconstructed image $\hat{x}$. The generator is trained by minimizing the weighted sum of $S_{con}$, $S_{enc}$ and $S_{adv}$, which have defined in Eq.\ref{eq:S_con}-\ref{eq:S_adv}~\cite{ganomaly:2018}. Therefore, the  distillation loss $K_l$ described in Eq.\ref{eq:K_l} is designed as the weighted sum of the losses $K_1$, $K_x$ and $K_2$, where $K_1$, $K_2$ represent the difference between the two latent vectors ($z_1$ and $\hat{z_1}$, $z_2$ and $\hat{z_2}$) in the teacher GAN and student GAN, and $K_x$ is the difference of two reconstructed images ($\hat{x}$, $\widetilde{x}$). 
	2) How to combine the distillation loss $K_l$ with existing generator losses $L_g^S, L_g^T$ and discriminator losses $L_d^S, L_d^T$ from the student and teacher GANs to improve the performance of the student GAN? As is illustrated in Figure \ref{fig:P-KDGAN}(b), we design four distillation structures (KDGAN-\normalsize{\textcircled{\footnotesize{1}}}, KDGAN-\normalsize{\textcircled{\footnotesize{2}}}, KDGAN-\normalsize{\textcircled{\footnotesize{3}}} and KDGAN-\normalsize{\textcircled{\footnotesize{4}}}) based on different combinations of the above five losses. The difference between them consists of two aspects: on the one hand, whether the weights of the teacher GAN are fixed; on the other hand, whether the distillation loss $K_l$ is combined with the losses $L_g^S, L_d^S$ of the student GAN for knowledge transfer. 
	3) Whether the designed four distillation structures can make the performance of the student GAN like that of the teacher GAN? If not, how to fix it? Our experimental results demonstrate that the performance of student GANs trained from scratch (or with random initialization) by the above four distillation structures is incomparable with the teacher GAN. Just as learning is a gradual cumulative process, a two-step progressive learning of KDGAN is proposed to continuously improve the performance of the student GAN. In the first step, the student GAN imitates the representation of the pre-trained teacher GAN with fixed weights to make itself have a certain ``knowledge'' reserve. Such a "teaching by teacher" step make the student learn the basic knowledge totally from the teacher. In the second step, the student GAN with basic ``knowledge'' reserve is fine-trained together with the teacher GAN. The second step of "fine-learning with teacher" can further improve the performance and stability by jointly utilizing the basic knowledge of the teacher and student.

	The performance of proposed progressive knowledge distillation with GANs for one-class novelty detection is evaluated on CIFAR-10~\cite{CIFAR10:2009}, MNIST~\cite{MNIST:2010} and FMNIST~\cite{FMNIST:2017} datasets. Our contributions are summarized as follows.
	\begin{itemize}
		\item We utilize the encoder-decoder-encoder based GAN for one-class novelty detection, which outperforms all the state-of-the-art methods.
		
		\item We propose new distillation losses on latent vectors and reconstructed images of GANs that allow the student to better learn from the teacher.
		
		\item  We regard the distillation process as a knowledgeable teacher to improve the performance and stability of student networks through two-step progressive learning, which includes basic knowledge learning and fine-learning.
		
		\item Progressive Knowledge Distillation with GANs is proposed for one-class novelty detection. Our experiments demonstrate that the P-KDAGN can improve the performance of the student GAN on the three datasets CIFAR-10, MNIST and FMNIST by 2.44\%, 1.77\%, and 1.73\%, respectively. 
	\end{itemize}

	\section{Related Work}
	We briefly review the related works in term of one-class novelty detection and knowledge distillation, as well as the architecture of Ganomaly~\cite{ganomaly:2018}.

	\subsection{One-class Novelty Detection}
	In the unsupervised one-class novelty detection, only the normal samples with one class are used for training the model. Conventionally, novelty detection methods can be divided into two categories~\cite{survey:2009}. One is the traditional methods, such as One-Class SVM (OC-SVM)~\cite{OCSVM:2001}, Kernel Density Estimation (KDE)~\cite{KDE:1962} and Principal Component Analysis (PCA)~\cite{PCA:1987}. The disadvantage of such approaches is that they are not suitable for high-dimensional image data. The other methods based on deep learning include Deep Belief Networks (DBN)~\cite{DBN:2016}, Autoencoders (AE)~\cite{DenoisAE:2008} and generative adversarial networks (GANs)~\cite{AnoGAN:2017,EfficientGAN:2018,OCGAN:2019}.
	
	GANs have shown state-of-the-art performance in modeling complex high-dimensional image distributions~\cite{GAN:2014}. Therefore, a lot of GANs based methods have been used for novelty detection~\cite{AnoGAN:2017,EfficientGAN:2018,OCGAN:2019}. The reconstruction errors of images or latent vectors are utilized as novelty score, which means that the learned model only reconstructs normal samples well, and shows very low tolerance for novel samples. Schlegl et al.~\cite{AnoGAN:2017} proposed the first GANs based work, AnoGAN, for novelty detection. In training, the combination of the residual loss on images and discrimination loss on feature maps is minimized to iteratively search the best latent vector. The Efficient GAN~\cite{EfficientGAN:2018} based on BiGAN~\cite{BiGAN:2016} network was proposed for jointly training the map from the image to the latent space simultaneously. Perera et al.~\cite{OCGAN:2019} proposed the OCGAN in which two discriminators were used in the latent space and the input space for making the learned network better model the input images. Recently, Ganomaly~\cite{ganomaly:2018} shown in Figure \ref{fig:P-KDGAN}(a) constructs a novel architecture for multi-class anomaly detection. In our method, the Ganomaly framework is used for one-class novelty detection.

	\subsection{Knowledge Distillation}
	To reduce the large computation and storage cost of deep convolutional neural networks, knowledge distillation can transfer the generalization ability of a large network (or an ensemble of networks) to a light-weight network. Hinton et al. ~\cite{Hinton:2015} used the outputs of the softmax layer of a teacher network as the target function to train the student network. Romero et al.~\cite{Fitnets:2015} proposed that a student network with random initialization can imitate the intermediate representations of the teacher network to improve its own performance. In order to ensure the student network to learn the true data distribution from the teacher network, knowledge distillation with a discriminator was used for distinguishing features extracted from the teacher and student networks~\cite{KDGAN:2018,Portable:2018,KTAN:2018}. In our method, knowledge distillation is considered as a progressive learning process, which can continuously improve the performance of student networks.
	
	GANs~\cite{GAN:2014} have been applied to many real world applications such as domain transfer, image generation, and novelty detection. However, to our knowledge, there is no related works that deploy the knowledge distillation on two standard GANs. Therefore, this paper designs a distillation loss to transfer knowledge from the teacher GAN to the student GAN.

	\subsection{The Architecture of Ganomaly}
	Akcay et al.~\cite{ganomaly:2018} proposed Ganomaly for multi-class anomaly detection, in which multiple class of samples as normal data and one class of samples as abnormal data. In this paper, we utilize Ganomaly architecture for one-class novelty detection. One-class means that only the instances in one category are regarded as normal data, and the remaining categories are abnormal data.
	
	GAN consists of two adversial modules, a generator $G$ and a discriminator $D$. As is shown in the student GAN of Figure \ref{fig:P-KDGAN}(a), the Ganomaly~\cite{ganomaly:2018} framework is composed of two modules: 1) an encoder-decoder-encoder ($G_E-G_D-G_R$) pipeline based generator $G$ that learns the distribution of input image $x$, where $x \in \mathbb{R}^{w\times h\times c}$, from latent spaces $z_1, z_2$, where $z_1, z_2 \in \mathbb{R}^{d}$; 2) a discriminator $D$ that decides whether the reconstructed image $\hat{x}$ is real or fake. $D$ and $G$ are simultaneously optimized by playing the following minmax game as:
	
	\begin{equation}
	\label{eq:GAN}
	\begin{split}
	\min_G\max_D V(D, G) =& \mathbb{E}_{x\sim X}[logD(x)] + \\
	+ & \mathbb{E}_{x\sim X}[log(1-G_D(G_E(x)))],
	\end{split}
	\end{equation}
	where training dataset $X$ comprises $N$ normal images, $X=[x_1, x_2, ..., x_N] \in \mathbb{R}^{N\times w \times h \times c}$, and $\mathbb{E}_{x\sim X}$ is the expected value of x obeying distribution of normal images $X$.
	
	During training, the generator loss $L_g^S$ and the cross-entropy loss $S_{ce}$ are minimized to train the student $G$ and student $D$, respectively. The model is trained from normal samples, therefore the reconstruction error is large on abnormal samples. In the previous methods~\cite{AnoGAN:2017,EfficientGAN:2018,OCGAN:2019}, the reconstruction errors of the images or latent vectors are used for anomaly detection. In Ganomaly~\cite{ganomaly:2018}, the difference $S_{enc}$ between two latent vectors $z_1$, $z_2$ is used as novelty score, which is defined in Eq. \ref{eq:S_enc}.

	\section{Our Method}
	In this work, we adopt the Ganomaly~\cite{ganomaly:2018} framework for one-class novelty detection and achieve state-of-the-art performance. To compress deep neural networks and deploy them to embedded devices with limited resources, we propose the Progressive Knowledge Distillation with GANs (P-KDGAN) to learn a lightweight student GAN from a pre-trained teacher GAN. The P-KDGAN method is composed of three modules. 1) Knowledge distillation with GANs (KDGAN), in which the distillation losses based on Ganomaly framework are proposed for transferring knowledge from the teacher GAN to the student GAN. 2) Four distillation structures are designed for KDGAN. 3) The two-step progressive learning of KDGAN can continuously improve the performance of the student GAN.

	\subsection{KDGAN}
	In the KDGAN, both the teacher GAN and the student GAN follow the network architecture of Ganomaly~\cite{ganomaly:2018} and have the same network layers. The difference between them is the number of channels in each layer. Therefore, the generator loss $L_g^T$ and the discriminator loss $L_d^T$ in the teacher GAN have the same form as $L_g^S$ and $L_d^S$. The generator loss $L_g^S$ of student GAN includes reconstructed image loss $S_{con}$, latent space loss $S_{enc}$ and adversarial loss $S_{adv}$:

	\begin{equation}
	\label{eq:S_con}
	S_{con} = \mathbb{E}_{x\sim X} \left\| x-\hat{x} \right\| _1,
	\end{equation}
	
	\begin{equation}
	\label{eq:S_enc}
	S_{enc} = \mathbb{E}_{x\sim X} \left\| z_1-z_2 \right\| _2, 
	\end{equation}
	
	\begin{equation}
	\label{eq:S_adv}
	S_{adv} = \mathbb{E}_{x\sim X} \left\| f(x)-f(\hat{x}) \right\| _2,
	\end{equation}
	
	\begin{equation}
	\label{eq:L_g^S}
	L_g^S = w_{con}S_{con} + w_{enc}S_{enc} + w_{adv}S_{adv},
	\end{equation}
	where $f(\cdot)$ outputs the intermediate representations of discriminator $D$. $S_{con}$, $S_{enc}$ and $S_{adv}$ denote the reconstruction errors of the images, latent vectors and feature maps, respectively. The weighted sum of $S_{con}$, $S_{enc}$ and $S_{adv}$ constitutes the generator loss $L_g^S$ and is minimized to train the student generator $G$. The discriminator loss $L_d^T$ consists of $S_{ce}$.
	
	The designed distillation loss $K_l$ is a novel attempt for knowledge distillation on two standard GANs. As is shown in Figure 1(a), the teacher GAN and the student GAN transfer knowledge through the intermediate layers of the generators, which includes two latent vectors and one reconstructed images. In the KDGAN, we design three losses $K_1$, $K_x$ and $K_2$ to measure the similarity of the intermediate layers. $K_1$ and $K_2$ are the $L2$ distance of latent vectors ($z_1$ and $\hat{z_1}$, $z_2$ and $\hat{z_2}$) from the teacher GAN and student GAN. $K_x$ is the $L1$ distance of reconstructed images ($\hat{x}$, $\widetilde{x}$). Based on the above three losses, we propose distillation loss $K_l$ as an objective function for knowledge distillation which is the weighted sum of $K_1$, $K_x$ and $K_2$:   
	
	\begin{equation}
	\label{eq:K_l}
	K_l = w_{1}K_{1} + w_{x}K_{x} + w_{2}K_{2},
	\end{equation}

	\subsection{Distillation Structures}
	\label{sec:3.2}
	As is shown in Figure \ref{fig:P-KDGAN}(a), the designed distillation loss $K_l$ builds a "bridge" between the teacher and student GANs for knowledge transfer. The losses in the KDGAN consist of three parts: teacher GAN losses $L_g^T$, $L_d^T$, student GAN losses $L_g^S$, $L_d^S$, and distillation loss $K_l$.  We define the above five loss functions as the elements of set $\mathcal{L}$:
	
	\begin{equation}
	\label{eq:L}
	\mathcal{L} = \left\{ \alpha L_g^T, \beta L_d^T, \mu L_g^S, \nu L_d^S, \lambda K_l \right\},
	\end{equation}
	where $\alpha$, $\beta$, $\mu$, $\nu$, and $\lambda$ $\in$ $\left\{0, 1 \right\}$ indicates whether the corresponding loss is used to train the networks.

	The elements in $\mathcal{L}$ can be combined into four subsets ($\mathcal{L}_1$, $\mathcal{L}_2$, $\mathcal{L}_3$, $\mathcal{L}_4$) to form different distillation structures according to the following two rules. The first rule is whether the teacher GAN has fixed weights; the second rule is whether the distillation loss $K_l$ is combined with the losses $L_g^S$, $L_d^S$ to train the student GAN. Before the KDGAN, a teacher GAN is trained by its own generator loss $L_g^T$ and discriminator loss $L_d^T$. The designed four distillation structures are introduced as follows:

	\begin{itemize}
		\item \textbf{KDGAN-\normalsize{\textcircled{\footnotesize{1}}}:} $\mathcal{L}_1$=$\left\{ K_l \right\}$. Without the use of real labels, the training of student network only depends on the distillation loss $K_l$, which results in poor detection performance. There is no adversarial networks, and the teacher network is not updated, so its training speed is the fastest. 
		
		\item \textbf{KDGAN-\normalsize{\textcircled{\footnotesize{2}}}:} $\mathcal{L}_2$=$\left\{ L_g^S, L_d^S, K_l \right\}$. The student GAN is trained by minimizing its own losses $L_g^S$, $L_d^S$ and distillation loss $K_l$, while the teacher GAN is not updated. The adversarial network in student GAN causes its training speed to be slightly slower than KDGAN-\normalsize{\textcircled{\footnotesize{1}}}.
		
		\item \textbf{KDGAN-\normalsize{\textcircled{\footnotesize{3}}}:} $\mathcal{L}_2$=$\left\{ L_g^T, L_d^T, K_l \right\}$. The teacher GAN uses its own losses $L_g^T$, $L_d^T$ to train to maintain its performance, when the training of the student GAN follows KDGAN-\normalsize{\textcircled{\footnotesize{1}}}. Its training speed is almost the same as that of KDGAN-\normalsize{\textcircled{\footnotesize{2}}}.
		
		\item \textbf{KDGAN-\normalsize{\textcircled{\footnotesize{4}}}:} $\mathcal{L}_2$=$\left\{ L_g^T, L_d^T, L_g^S, L_d^S, K_l \right\}$. The trainings of the teacher and student GANs follow KDGAN-\normalsize{\textcircled{\footnotesize{3}}} and KDGAN-\normalsize{\textcircled{\footnotesize{2}}}, respectively. There are two adversarial networks that need to be trained simultaneously, so the training speed is the slowest.
	\end{itemize}

	\begin{algorithm}[tb]
		\caption{Progressive Knowledge Distillation with GANs}
		\label{alg:algorithm}
		
		\textbf{Input}: Pre-trained teacher $G$ and $D$, training dataset with normal instances ${{(x_i,y_i)}}_{i=1}^{N}$, epoch $M$  \\
		\textbf{Output}: Improved student $G$
		
		\begin{algorithmic}[1] 
			\STATE \textbf{First step }
			\STATE{Student $G$ and $D$ with random initialization}
			
			\FOR{$m$=1 to $M$}
			\FOR{$i$=1 to $N$}
			\STATE{Update student GAN when teacher GAN with fixed weights}
			\ENDFOR
			\ENDFOR
			
			\STATE \textbf{Second step}
			\STATE{Download the weights of the teacher and student GANs from the previous step}
			\FOR{$m$=1 to $M$}
			\FOR{$i$=1 to $N$}
			\STATE{Update student GAN and teacher GAN together}
			\ENDFOR
			\ENDFOR
			
		\end{algorithmic}
	\end{algorithm}

	\subsection{Progressive Learning of KDGAN}
	The progressive learning of KDGAN, shown in Figure \ref{fig:P-KDGAN}(b), is a two-step approach that continuously improves the performance of the student GAN and achieves better performance than the single step methods. The two-step P-KDGAN is described as follows.
	
	\paragraph{P-KDGAN-I.}
	In the first step, four distillation structures are utilized to train student network. The experimental results shown in Section \ref{sec:4.4} demonstrate that the performance of student network with random initialization has a large gap compared with teacher network. Therefore, considering the detection accuracy and training time of the four distillation structures, KDGAN-\normalsize{\textcircled{\footnotesize{2}}} is used as the first step of P-KDGAN to enable student network to learn the basics knowledge from teacher network. In the KDGAN-\normalsize{\textcircled{\footnotesize{2}}}, the pre-trained teacher has already converged, so the teacher network with fixed weights is used to train the student network relying on real labels and distillation knowledge.

	\paragraph{P-KDGAN-II.}
	In the second step, KDGAN-\normalsize{\textcircled{\footnotesize{3}}} and KDGAN-\normalsize{\textcircled{\footnotesize{4}}} continue to train the teacher networks, while the student networks with basic knowledge rely on distilling knowledge to fine-training, thereby further improving accuracy and stability. The fine-learning processes in this step are named as P-KDGAN-II-\normalsize{\textcircled{\footnotesize{2}}}\normalsize{\textcircled{\footnotesize{3}}} and P-KDGAN-II-\normalsize{\textcircled{\footnotesize{2}}}\normalsize{\textcircled{\footnotesize{4}}}. The experimental results prove that the performance of student network even exceeds the teacher network in some categories of one-class novelty detection.
	
	The above process is illustrated in Algorithm \ref{alg:algorithm}.

	\section{Experiments}
	In this section, the proposed P-KDGAN is evaluated on the well-known CIFAR-10~\cite{CIFAR10:2009}, MNIST~\cite{MNIST:2010} and FMNIST~\cite{FMNIST:2017} datasets. Following previous work~\cite{OCGAN:2019}, we quantify the performance of our method using the Area Under Curve (AUC) of Receiver Operating Characteristics (ROC). The performance results are analyzed in details and are compared with state-of-the-art techniques.
	
	All the reported results are implemented using the PyTorch framework~\cite{pytorch} on NVIDIA TITAN 2080Ti. In the experiments, the batch size and epoch are set to 1 and 500 respectively. Adam~\cite{adam:2015} is used for training with a learning rate of 0.002.
	
	\begin{figure*}
		\centering
		\subfigure[]{
			\begin{minipage}{0.325\linewidth}
				\centering
				\includegraphics[scale=0.418]{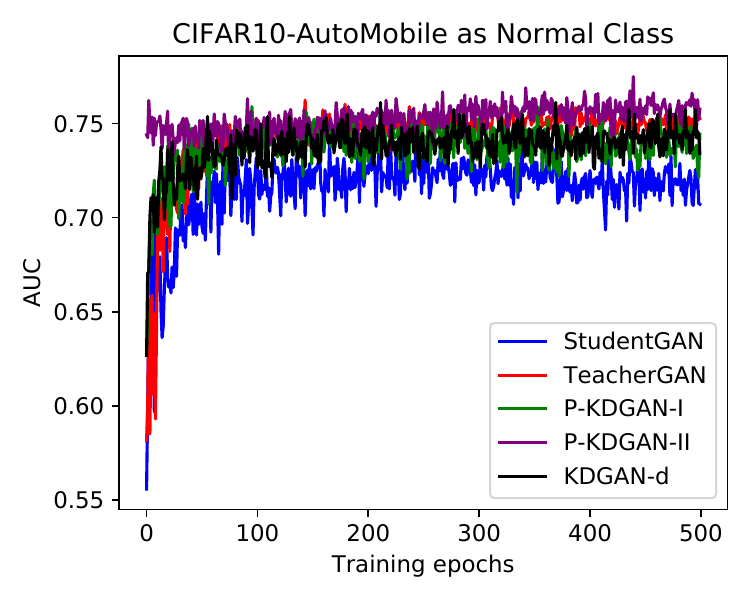}
			\end{minipage}%
		}%
		\subfigure[]{
			\begin{minipage}{0.325\linewidth}
				\centering
				\includegraphics[scale=0.418]{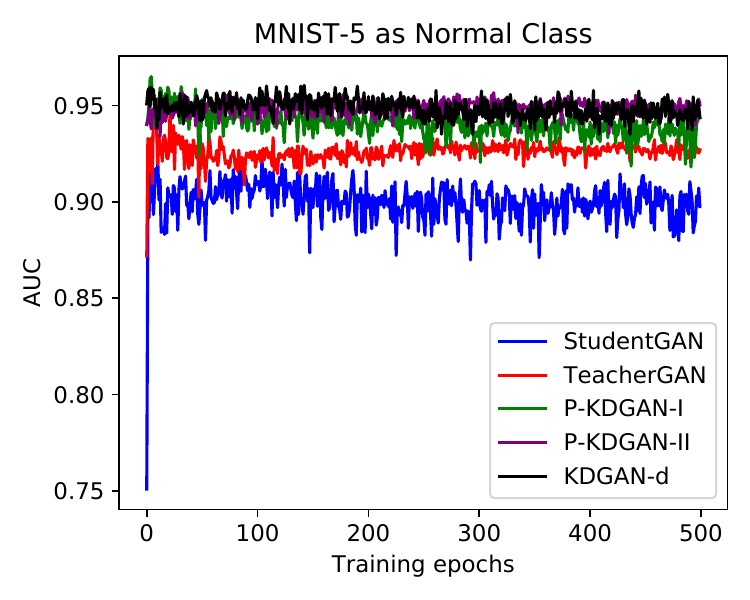}
			\end{minipage}
		}%
		\subfigure[]{
			\begin{minipage}{0.325\linewidth}
				\centering
				\includegraphics[scale=0.418]{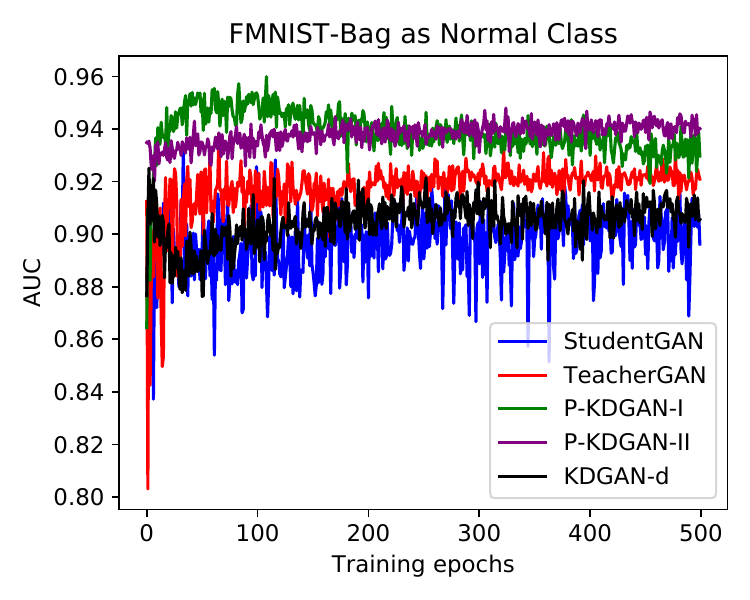}
			\end{minipage}%
		}%
		
		\centering
		\caption{Training curves of the AUC on three datasets. The normal classes are: (a) AutoMobile on CIFAR-10. (b) 5 on MNIST. (c) Bag on FMNIST. P-KDGAN-I represents KDGAN-\normalsize{\textcircled{\footnotesize{2}}}. P-KDGAN-II represents P-KDGAN-II-\normalsize{\textcircled{\footnotesize{2}}}\normalsize{\textcircled{\footnotesize{3}}}. KDGAN-d represents KDGAN-\normalsize{\textcircled{\footnotesize{4}}}.}\label{fig:KDGAN-Vs.P-KDGAN}
	\end{figure*}

	\subsection{Datasets}
	For the three experimental datasets, the training and testing partitions remain as default. In the setup, one of the classes from training dataset is considered as normal samples for training. During testing, the remaining classes are used to represent novelty samples. For example, every experiment on the CIFAR-10 dataset is trained with 5000 samples and tested with 10,000 samples. The above experiment is repeated for all the ten categories. In addition, in order to compatible with the network architectures, all the images are resized to 32$\times$32 by Bilinear interpolation.
	
	\begin{table}
		\centering
		\scalebox{0.625}{
			\begin{tabular}{lcccc}  
				\toprule
				Layer      & Units  & BN            & Activation   & Kernel           \\
				\midrule
				$E(\textit x)$                                                          \\
				Conv2D     & 64     & $\checkmark$  & LeakyReLU        & $4\times4$   \\
				Conv2D     & 128    & $\checkmark$  & LeakyReLU        & $4\times4$   \\
				Conv2D     & 256    & $\checkmark$  & LeakyReLU        & $4\times4$   \\
				Conv2D     & 256    &               &                  & $4\times4$   \\
				
				$D(\textit x)$                                                          \\
				ConvTrans2D     & 256     & $\checkmark$  & ReLU       & $4\times4$   \\
				ConvTrans2D     & 128     & $\checkmark$  & ReLU       & $4\times4$   \\
				ConvTrans2D     & 64      & $\checkmark$  & ReLU       & $4\times4$   \\
				ConvTrans2D     & 3       &               & Tanh       & $4\times4$   \\
				\bottomrule
		\end{tabular}}
		\caption{The encoder and decoder architecture for our teacher GAN, layer by layer. Units refer to number of filters in the case of convolution layers, and BN is Batch Normalization abbreviated.}\label{tab:archtecture}
	\end{table}

	\subsection{Network Architectures}
	The Ganomaly~\cite{ganomaly:2018} framework based on encoder-decoder-encoder ($G_E-G_D-G_R$) pipeline is used in our method. $G_E$, $G_R$ in the generator and discriminator $D$ are encoders, $G_D$ is decoder. The encoder $E(x)$ and decoder $D(x)$ follow the DCGAN~\cite{DCGAN:2015} architecture, which have three basic layers in our model. As is shown in Table 1, the basic layers consist of: convolutional layers (deconvolutional layers), batch normalization and activation. In contrast, LeakyReLU and ReLU activations are used in encoders and decoders, except for the last layer in decoder, which uses Tanh. All the convolution filters are set to $4\times4$. 
	
	The difference between a teacher network and a student network is the number of channels in the intermediate representations. For the three experimental datasets, the intermediate layers in the teacher networks are set to 64-128-256 channels following the OCGAN~\cite{OCGAN:2019}. The student networks in each dataset utilize intermediate representations with 8-16-64 channels, 2-4-8 channels and 1-2-4 channels respectively. The encoder $E(x)$ and decoder $D(x)$ architecture of the teacher GAN is illustrated in Table \ref{tab:archtecture}.
	
	\begin{table}
		\centering
		\vspace{1mm}
		\scalebox{0.54}{
			\renewcommand{\arraystretch}{1.1}
			\begin{tabular}{l|cccccccc}  
				\hline
				NORMAL CLASS & OCSVM & KDE    & VAE    & AND    & AnoGAN & DSVDD  & OCGAN  & Ours \\
				
				\hline
				AIRPLANE    & 0.630  & 0.658  & 0.700  & 0.717  & 0.671  & 0.617  & 0.757  & \textbf{0.825} \\
				
				AUTOMOBILE  & 0.440  & 0.520  & 0.386  & 0.494  & 0.547  & 0.659  & 0.531  & \textbf{0.744} \\  
				
				BIRD        & 0.649  & 0.657  & 0.679  & 0.662  & 0.529  & 0.508  & 0.640  & \textbf{0.703} \\
				
				CAT         & 0.487  & 0.497  & 0.535  & 0.527  & 0.545  & 0.591  & \textbf{0.620}  & 0.605 \\
				
				DEER        & 0.735  & 0.727  & 0.748  & 0.736  & 0.651  & 0.609  & 0.723  & \textbf{0.765} \\
				
				DOG         & 0.500  & 0.496  & 0.523  & 0.504  & 0.603  & \textbf{0.657}  & 0.620  & 0.652 \\
				
				FROG        & 0.725  & 0.758  & 0.687  & 0.726  & 0.585  & 0.677  & 0.723  & \textbf{0.797} \\
				
				HORSE       & 0.533  & 0.564  & 0.493  & 0.560  & 0.625  & 0.673  & 0.575  & \textbf{0.723} \\
				
				SHIP        & 0.649  & 0.680  & 0.696  & 0.680  & 0.758  & 0.759  & 0.820  & \textbf{0.827} \\
				
				TRUCK       & 0.508  & 0.540  & 0.386  & 0.566  & 0.665  & 0.731  & 0.554  & \textbf{0.735} \\	
				\hline
				MEAN        & 0.5856 & 0.6097 & 0.5833 & 0.6172 & 0.6179 & 0.6481 & 0.6566 & \textbf{0.7376} \\
				\hline
				
				0           & 0.988  & 0.885  & 0.997  & 0.984  & 0.966  & 0.980  & \textbf{0.998}  & 0.996 \\
				
				1           & 0.999  & 0.996  & 0.999  & 0.995  & 0.992  & 0.997  & \textbf{0.999}  & \textbf{0.999} \\  
				
				2           & 0.902  & 0.710  & 0.936  & 0.947  & 0.850  & 0.917  & 0.942  & \textbf{0.969} \\
				
				3           & 0.950  & 0.693  & 0.959  & 0.952  & 0.887  & 0.919  & 0.963  & \textbf{0.969} \\
				
				4           & 0.955  & 0.844  & 0.973  & 0.960  & 0.894  & 0.949  & \textbf{0.975}  & 0.970 \\
				
				5           & 0.968  & 0.776  & 0.964  & 0.971  & 0.883  & 0.885  & \textbf{0.980}  & 0.951 \\
				
				6           & 0.978  & 0.861  & 0.993  & 0.991  & 0.947  & 0.983  & 0.991  & \textbf{0.992} \\
				
				7           & 0.965  & 0.884  & 0.976  & 0.970  & 0.935  & 0.946  & 0.981  & \textbf{0.982} \\
				
				8           & 0.853  & 0.669  & 0.923  & 0.922  & 0.849  & 0.939  & 0.939  & \textbf{0.965} \\
				
				9           & 0.955  & 0.825  & 0.976  & 0.979  & 0.924  & 0.965  & 0.981  & \textbf{0.987} \\	
				\hline
				MEAN        & 0.9513 & 0.8143 & 0.9696 & 0.9671 & 9127   & 0.9480 & 0.9750 & \textbf{0.9780}  \\
				\hline
		\end{tabular}}
		\caption{One-class novelty detection results on CIFAR-10 and MNIST dataset. The average AUC of three repeated experiments was used as detection performance.}\label{tab:Results}
	\end{table}

	\subsection{Results on One-class Novelty Detection}
	In this section, we compare our Ganomaly~\cite{ganomaly:2018} based Teacher GAN with several traditional and deep learning based methods on CIFAR-10 and MNIST datasets , including one-class SVM (OC-SVM)~\cite{OCSVM:2001}, kernel density estimation (KDE)~\cite{KDE:1962}, deep variational autoencoder (VAE)~\cite{VAE:2013}, AND~\cite{AND:2019}, AnoGAN~\cite{AnoGAN:2017}, DSVDD~\cite{DSVDD:2018} and OCGAN~\cite{OCGAN:2019}. In light of massive experiments, the parameters of $w_{con}$, $w_{enc}$ and $w_{adv}$ in Eq. \ref{eq:L_g^S} are mannually configured as 10, 1 and 1. The parameters of $w_1$, $w_x$ and $w_2$ in Eq. \ref{eq:K_l} are set as 1. We take the average AUC of the last epoch from multiple trials, but not the manually selected result, as the detection performance, which is more convictive.
	
	\paragraph{Comparisons on CIFAR-10 and MNIST.} The performance of one-class novelty detection on CIFAR-10 dataset, our method shown in Table \ref{tab:Results} achieves 0.7376, which is higher than the best OCGAN~\cite{OCGAN:2019} method about 8\%. For MNIST dataset, our method achieves 0.978 yielding an improvement of about 0.3\% compared with state-of-the-art method.

	\subsection{Evaluation of P-KDGAN Method}
	\label{sec:4.4}
	In this section, the progressive knowledge distillation with GANs is evaluated on CIFAR-10, MNIST and FMNIST datasets. In each experiment, the weights of the last epoch are served as the teacher network.
	
	\begin{table}
		\centering
		\scalebox{0.7}{
			\renewcommand{\arraystretch}{1}
			\begin{tabular}{cccc}  
				\toprule
				Method                                              & CIFAR-10   & MNIST                    & FMNIST                       \\
				
				\midrule 
				KDGAN-\normalsize{\textcircled{\footnotesize{1}}}   & 0.7159    & 0.9670    & 0.9248                                     \\

				KDGAN-\normalsize{\textcircled{\footnotesize{2}}}   & 0.7243    & 0.9675   & 0.9241                                    \\

				KDGAN-\normalsize{\textcircled{\footnotesize{3}}}   & 0.7094    & {\color{blue} 0.9718}      & {\color{blue} 0.9290}   \\

				KDGAN-\normalsize{\textcircled{\footnotesize{4}}}   & {\color{blue} 0.7258}       & 0.9687   & 0.9242               \\
				
				\midrule
				P-KDGAN-II-\normalsize{\textcircled{\footnotesize{2}}}\normalsize{\textcircled{\footnotesize{3}}}  & {\color{red} 0.7305}    & {\color{red} 0.9725}   & {\color{red} 0.9293}  \\

				P-KDGAN-II-\normalsize{\textcircled{\footnotesize{2}}}\normalsize{\textcircled{\footnotesize{4}}}  & 0.7252     & 0.9667    & 0.9257                             \\
				
				\bottomrule
		\end{tabular}}
		\caption{Compare the performance of KDGAN and P-KDGAN. We highlight the best results in {\color{red} red} and the second-best results in {\color{blue} blue} color.}\label{tab:KDGAN-Vs.P-KDGAN}
	\end{table}

	\paragraph{KDGAN vs. P-KDGAN.} As is shown in Table \ref{tab:KDGAN-Vs.P-KDGAN}, P-KDGAN-II-\normalsize{\textcircled{\footnotesize{2}}}\normalsize{\textcircled{\footnotesize{3}}} achieves the best performance on three datasets, which illustrates the effectiveness of our progressive learning of KDGAN. Although KDGAN-\normalsize{\textcircled{\footnotesize{3}}} achieves the second-best results on MNIST and FMNIST,  it shows the worst performance on CIFAR-10 dataset. KDGAN-\normalsize{\textcircled{\footnotesize{4}}} obtain the second-best results on CIFAR-10, but it was about 0.5\% lower than the best result. In addition, KDGAN-d (KDGAN-\normalsize{\textcircled{\footnotesize{4}}}) iillustrated in Figure \ref{fig:KDGAN-Vs.P-KDGAN}(c) is inferior in accuracy and training stability compared to P-KDGAN-II. The training curves of the AUC illustrated in Figure \ref{fig:KDGAN-Vs.P-KDGAN} clearly shows that proposed P-KDGAN-II can improve the accuracy of the student network and even surpass the teacher network, and reduce shock. Therefore, the above analysis concludes that student networks with random initialization can only learn the basic knowledge of the teacher networks, and the fine-training in the second step of P-KDGAN can further improve performance.
	
	\begin{table}
		\centering
		\scalebox{0.565}{
			\renewcommand{\arraystretch}{1}
			\begin{tabular}{ccccccc}  
				\toprule
				Dataset  & Method & AUC. $\downarrow$ & \#Param. $\downarrow$ & \#FLOPs. $\downarrow$ \\
				
				\midrule
				\multirow{3}{*}{CIFAR-10} 	& Teacher    & 0.7376   & 5.12M           & 56M                      \\
				\cmidrule{2-5}
				& Student    & 3.15\%            & \multirow{2}{*}{6.22$\times$}    & \multirow{2}{*}{24.45$\times$}            \\
				& P-KDGAN    & \textbf{0.71\%}   &                       &    \\
				
				\midrule
				\multirow{3}{*}{MNIST}  & Teacher  & 0.9780   & 5.12M           & 56M                      \\
				\cmidrule{2-5}
				& Student    & 2.32\%              & \multirow{2}{*}{52.22$\times$}   & \multirow{2}{*}{311.11$\times$}           \\
				& P-KDGAN    & \textbf{0.55\%}     &           &   \\
				
				\midrule
				\multirow{3}{*}{FMNIST} & Teacher    & 0.9311    & 5.12M           & 56M                   \\
	            \cmidrule{2-5}
				& Student     & 1.91\%               & \multirow{2}{*}{105.45$\times$}  & \multirow{2}{*}{700$\times$}          \\
				& P-KDGAN     & \textbf{0.18\%}      &   &  \\
				
				\bottomrule
		\end{tabular}}
		\caption{Evaluation of our P-KDGAN method on CIFAR-10, MNIST and FMNIST datasets. (M means million, $\#$ means the compression ratio of parameter numbers and FLOPs compared to the teacher GAN.)}
		\label{tab:compression}
	\end{table}

	\paragraph{Results on P-KDGAN.} As is illustrated in Table \ref{tab:compression}, the performance of the student GAN obtained by two-step P-KDGAN is only 0.71\%, 0.55\% and 0.18\% lower than that of the teacher GAN when compressing the computation at ratios of 24.45:1, 311.11:1, and 700:1, respectively.

	\section{Conclusion}
	In this paper, we use the encoder-decoder-encoder pipeline based GANs for one-class novelty detection and achieve state-of-the-art performance. To compress the model, the progressive knowledge distillation with GANs is proposed, which is a novel exploration that applies the knowledge distillation on two standard GANs. The two-step progressive learning can continuously improve the performance and reduce shock of the student network, in which the designed distillation loss plays an important role. Experiments on three datasets validate the effectiveness of our proposed method. Moreover, our proposed method can be used to compress other GANs-based applications, such as image generation.
	
	\section*{Acknowledgments}
	This work is supported by Key-Area Research and Development Program of Guangdong Province (2019B010155003), National Natural Science Foundation of China (U1713203), Shenzhen Science and Technology Innovation Commission (Project KQJSCX20180330170238897), and the Scientific  Instrument  Developing Project of the Chinese Academy of Sciences (Grant No. YJKYYQ20190028).

	\bibliographystyle{ijcai20}
	\bibliography{ijcai20}
	
	
\end{document}